\documentclass[10pt,journal,compsoc]{IEEEtran}
\ifCLASSOPTIONcompsoc
  \usepackage[nocompress]{cite}
\else
  \usepackage{cite}
\fi
\ifCLASSINFOpdf
\else
\fi
\usepackage{amsmath,amsfonts}

\usepackage{algorithm}
\usepackage{algorithmic}
\usepackage{graphicx}
\usepackage{textcomp}
\usepackage{xcolor}
\usepackage{pifont}
\usepackage{xspace}
\usepackage{soul}
\usepackage{url}
\usepackage{booktabs}
\usepackage{multicol}
\usepackage{multirow}
\usepackage{dcolumn}
\usepackage{algorithm}
\usepackage{threeparttable}
\usepackage{hyperref}

\usepackage{dsfont,multirow}
\usepackage{subfig}
\usepackage[group-separator={,},group-minimum-digits={3}]{siunitx}
\usepackage{amsmath,amsfonts,bm}

\def\eqref#1{equation~\ref{#1}}

\def\1{\bm{1}}

\def\ru{{\textnormal{u}}}
\def\rv{{\textnormal{v}}}

\def\rx{{\textnormal{x}}}

\def\rvv{{\mathbf{v}}}

\def\rvx{{\mathbf{x}}}

\def\ervv{{\textnormal{v}}}

\def\ervx{{\textnormal{x}}}

\def\rmY{{\mathbf{Y}}}

\def\mA{{\bm{A}}}

\def\mD{{\bm{D}}}

\def\mH{{\bm{H}}}

\def\mL{{\bm{L}}}

\def\mX{{\bm{X}}}
\def\mY{{\bm{Y}}}

\DeclareMathAlphabet{\mathsfit}{\encodingdefault}{\sfdefault}{m}{sl}
\SetMathAlphabet{\mathsfit}{bold}{\encodingdefault}{\sfdefault}{bx}{n}

\def\gG{{\mathcal{G}}}

\newcommand{\E}{\mathbb{E}}

\newcommand{\KL}{D_{\mathrm{KL}}}

\newcommand{\ours}{\textsc{ReDiSC}\xspace}
\newtheorem{theorem}{Theorem}

\newtheorem{lemma}{Lemma}
\newtheorem{proposition}[theorem]{Proposition}

\begin{document}

\title{ReDiSC: A Reparameterized Masked Diffusion Model for Scalable Node Classification with Structured Predictions}
\author{
    Yule Li, 
    Yifeng Lu, 
    Zhen Wang$^{\ast}$,
    Zhewei Wei,
    Yaliang Li, 
    Bolin Ding

    \thanks{$^{\ast}$ Zhen Wang is the corresponding author.}
    \thanks{$\bullet$ Yule Li, Yifeng Lu, and Zhen Wang are with Sun Yat-sen University. Email: \{liyle3@mail2,luyf55@mail2,wangzh665@mail\}.sysu.edu.cn}
    \thanks{$\bullet$ Zhewei Wei is with Renmin University of China. Email: zhewei@ruc.edu.cn}
    \thanks{$\bullet$ Yaliang Li and Bolin Ding are with Alibaba Group. Email: \{yaliang.li,bolin.ding\}@alibaba-inc.com}

}

\IEEEtitleabstractindextext{%
\begin{abstract}
In recent years, graph neural networks (GNN) have achieved unprecedented successes in node classification tasks.
Although GNNs inherently encode specific inductive biases (e.g., acting as low-pass or high-pass filters), most existing methods implicitly assume conditional independence among node labels in their optimization objectives.
While this assumption is suitable for traditional classification tasks such as image recognition, it contradicts the intuitive observation that node labels in graphs remain correlated, even after conditioning on the graph structure.
To make structured predictions for node labels, we propose \textbf{ReDiSC}, namely, \textbf{Re}parameterized masked \textbf{Di}ffusion model for \textbf{S}tructured node \textbf{C}lassification.
\ours estimates the joint distribution of node labels using a reparameterized masked diffusion model, which is learned through the variational expectation-maximization (EM) framework.
Our theoretical analysis shows the efficiency advantage of \ours in the E-step compared to DPM-SNC, a state-of-the-art model that relies on a manifold-constrained diffusion model in continuous domain.
Meanwhile, we explicitly link \ours's M-step objective to popular GNN and label propagation hybrid approaches.
Extensive experiments demonstrate that \ours achieves superior or highly competitive performance compared to state-of-the-art GNN, label propagation, and diffusion-based baselines across both homophilic and heterophilic graphs of varying sizes. Notably, \ours scales effectively to large-scale datasets on which previous structured diffusion methods fail due to computational constraints, highlighting its significant practical advantage in structured node classification tasks.
The code for \ours is available \href{https://github.com/liyle3/REDISC}{here}.
\end{abstract}

}

\maketitle

\renewcommand*{\thefootnote}{\arabic{footnote}}

\IEEEdisplaynontitleabstractindextext

%
\IEEEpeerreviewmaketitle



\section{Introduction}
\label{sec:intro}
Node classification is a fundamental task in graph machine learning with numerous real-world applications, including social network analysis~\cite{social_network} and citation network modeling~\cite{citation1}. In this work, we focus on the transductive setting, where both labeled and unlabeled nodes—including those whose labels are to be predicted—are accessible during training. This setting differs from conventional classification tasks over i.i.d. (feature, label) pairs, as node labels in a graph are often interdependent even after conditioning on all node features. As a result, the central challenge lies in performing \textit{structured prediction} by fully leveraging node features, graph connectivity, and observed labels.

Graph Neural Networks (GNNs)~\cite{GCN, GraphSAGE, GAT, GIN} have become the leading approach for node classification, iteratively aggregating messages from each node's neighborhood to capture both attribute and structural information.
However, GNN-based methods often assume conditional independence among node labels in their optimization objective (see Sec.~\ref{sec:pre}).
This assumption may be inadequate for graphs with complex or long-range label dependencies, where inter-label correlations play a central role.

Label propagation (LP) methods offer an alternative to GNNs by treating transductive node classification as a label diffusion process based purely on graph topology~\cite{LP, TLLT, TPN}. These methods naturally support structured prediction and are highly efficient, but lack expressiveness: they cannot incorporate node features or learn task-specific propagation strategies.
To overcome the respective limitations of GNN and LP, several hybrid approaches have been proposed~\cite{PTA, LPA, parallel_propagate,LabelReuse}.
Despite promising results, many of them still rely on a factorized joint distribution over node labels, limiting their ability to fully capture label dependencies.


Recently, diffusion-based models such as DPM-SNC~\cite{DPM-SNC} and LGD~\cite{LGD} have been proposed for node classification by modeling the joint distribution of node labels via diffusion processes. These models generate predictions by repeatedly applying a GNN-based denoiser, enabling them to capture complex and long-range label dependencies. As a result, they achieve improved performance and consistency (as measured by subgraph-level accuracy; see Sec.~\ref{sec:exp}) over vanilla GNNs, LP methods, and traditional structured prediction models~\cite{GMNN, CL, SPN}. Despite these advantages, existing diffusion-based methods face several limitations:
(1) \textit{Scalability}: Predictions rely on manifold-constrained sampling or denoising in high-dimensional latent spaces, both of which incur substantial computational overhead and limit scalability.
(2) \textit{Semantic mismatch}: These models operate in continuous spaces, which are inherently misaligned with the discrete nature of classification tasks.
(3) \textit{Lack of interpretability}: They offer no explicit connections to classical label propagation methods, reducing transparency and insight into their behavior.

To overcome these limitations, we propose the \textbf{Re}parameterized masked \textbf{Di}ffusion model for \textbf{S}tructured \textbf{C}lassification (\textbf{ReDiSC}), specifically tailored for scalable node classification tasks.
\ours explicitly models the joint distribution of node labels using a reparameterized masked diffusion model (RMDM), which naturally aligns with the discrete nature of classification tasks, thereby addressing the semantic mismatch inherent in existing continuous-domain diffusion models.
Since only a minor fraction of nodes are labeled, we learn this joint distribution with the variational expectation-maximization (EM) framework, where we employ a priority queue to select more helpful samples from the variational distributions.
To resolve the scalability issue, the RMDM enables an efficient labeled-first inference procedure to solve the inverse problem of interest, significantly reducing computational overhead compared to existing manifold-constrained sampling methods.
Furthermore, each step of RMDM's reverse process corresponds to denoising the node labels based on all node features and part of node labels, which achieves interpretability by establishing a clear connection to the popular ``GNN+LP'' hybrids.
As the denoiser of RMDM carries out label denoising at various noise levels, we design a novel time-aware GNN layer for it.

Our main contributions are as follows:
(1) We propose \ours, the first discrete-domain diffusion model specifically designed for structured node classification, which effectively captures label dependencies comparable to continuous diffusion approaches while substantially reducing their computational overhead.
(2) We provide a theoretical comparison of \ours and recent diffusion-based models in terms of structured prediction capacity and time complexity. In addition, we show that \ours can be interpreted as an ensemble of ``GNN+LP'' hybrids that propagate both features and labels.
(3) Extensive experiments demonstrate that \ours generally outperforms state-of-the-art GNNs, label propagation, and recent diffusion-based baselines on both homophilic and heterophilic node classification benchmarks of varying scales, confirming its advantages in accuracy and scalability. Our ablation studies further validate the effectiveness of each key component, which can also benefit related diffusion-based models.
\section{Preliminaries}
\label{sec:pre}

\noindent\textbf{Node Classification.}
We consider an undirected graph $\gG=(\mathcal{V},\mathcal{E})$, where \(\mathcal{V}\) and \(\mathcal{E}\) denote the sets of nodes and edges, respectively.
Suppose $N=|\mathcal{V}|$, the node feature matrix is denoted by \(\mX \in \Re^{N \times d}\) with \( \mX_i \) representing the $d$-dimensional feature vector of the $i$-th node.
The adjacency matrix \( \mA \in \{0, 1\}^{N \times N}\) satisfies \( \mA_{ij} = 1 \) if \((i,j) \in \mathcal{E}\) and \( \mA_{ij} = 0 \) otherwise.
Each node belongs to one of the $C$ classes $\mathcal{Y}=\{1,\ldots,C\}$.
However, only a fraction of the nodes have been labeled while the remaining nodes' labels are unknown.
Thus, we use $\mathcal{V}_{\mathrm{L}}$ and $\mathcal{V}_{\mathrm{L}} = \mathcal{V}\setminus\mathcal{V}_{\mathrm{U}}$ to denote them, respectively.
Then, the node labels can be represented by \( \mY \in \Re^{N \times C}\), where each \(\mY_i\) is a $C$-dimensional one-hot vector denoting the label of the $i$-th node.
The labels of $\mathcal{V}_{\mathrm{L}}$ and $\mathcal{V}_{\mathrm{U}}$ are denoted by $\mY_{\mathrm{L}}$ and $\mY_{\mathrm{U}}$, respectively, although only $\mY_{\mathrm{L}}$ is observed~\footnote{We use $\mY_{\mathrm{L}}$ to denote either the submatrix of $\mY$ consisting of rows indexed by $\mathcal{V}_{\mathrm{L}}$ or the full $N\times C$ matrix with zero vectors in rows corresponding to $\mathcal{V}_{\mathrm{U}}$, depending on context.}.

\noindent\textbf{GNN-based Methods.}
To infer $\mY_{\mathrm{U}}$, most GNN-based methods simplify the estimated joint probability distribution over node labels as follows:
\begin{equation}
\begin{aligned}
    p_{\phi}(\rmY | \gG ) &:= \prod_{i\in\mathcal{V}} p_{\phi}(\rmY_i | \gG ),\\
    \forall i\in\mathcal{V}, p_{\phi}(\rmY_i | \gG ) &:= \hat{\mY}_{i}\text{ with }\hat{\mY} = g_{\phi}(\mX, \mA),
    \end{aligned}
    \label{eq:independence}
\end{equation}
where $\hat{\mY}\in [0,1]^{N \times C}$ denotes the soft labels predicted by a GNN model $g_{\phi}(\cdot,\cdot)$ with $\forall i,\sum_{j=1}^{C}\hat{\mY}_{i,j}=1$, and $\phi$ denotes this GNN's parameters.

With the independence assumption of Eq.~\ref{eq:independence}, this GNN model is often learned by maximizing the log-likelihood of observed $\mY_{\mathrm{L}}$ as follows:
\begin{equation}
\begin{aligned}
    \mathcal{L}_{\text{GNN}}(\phi)&=\log{ p_{\phi}(\rmY_{\mathrm{L}}=\mY_{\mathrm{L}} | \gG ) } = \log{ \prod_{i\in\mathcal{V}_{\mathrm{L}}} p_{\phi}( \rmY_{i} = \mY_{i} | \gG ) }\\
    &=\sum_{i\in\mathcal{V}_{\mathrm{L}}} \log{ p_{\phi}( \rmY_{i} = \mY_{i} | \gG ) } = -\sum_{i\in\mathcal{V}_{\mathrm{L}}}\mathrm{CE}(\mY_{i}, \hat{\mY}_i),
    \label{eq:gnnobj}
\end{aligned}
\end{equation}
where $\text{CE}(\cdot, \cdot)$ denotes cross-entropy.
Although Eq.~\ref{eq:gnnobj} is very common in classification tasks, node labels are usually correlated even after conditioning the graph, and relaxing the independence assumption to produce more structured predictions has been shown to be helpful~\cite{DPM-SNC}.

\noindent\textbf{Label propagation-based Methods.}
Since we focus on transductive node classification in this paper, label propagation is a widely adopted approach to produce structured predictions.
This kind of method~\cite{learning_consistency, iterative_algo} is often formulated as optimizing the predicted labels $\hat{\mY}$ by minimizing the objective:
$\mathcal{L}_{\text{LP}}(\hat{\mY}) = (1-\lambda)\|\hat{\mY}-\mY_{\mathrm{L}}\|_{2}^{2} + \lambda\text{tr}( \hat{\mY}^{\mathrm{T}}\mL \hat{\mY} )$,
where $\text{tr}(\cdot)$ means the trace of the input matrix, $\mL$ denotes the symmetric normalized Laplacian of $\gG$, and $\lambda\in(0, 1)$ balances the label alignment and smoothness constraints.

However, $\mathcal{L}_{\text{LP}}(\cdot)$ ignores the node features and relies on the smoothness assumption, which might not hold perfectly in practice~\cite{GPRGNN}.
Hence, a popular extension~\cite{parallel_propagate} is to feed the concatenation of the node features and node labels to a GNN for prediction: $\hat{\mY} = g_{\phi}( \mX \| \mY_{\mathrm{L}}, \mA)$.
To train this \textit{``GNN+LP'' hybrid}, label trick~\cite{LabelReuse} is adopted, and $\phi$ is learned by maximizing:
\begin{equation}
\begin{aligned}
\mathcal{L}_{\text{GNN+LP}}^{\lambda}(\phi) &= \E_{\mathcal{V}_{\text{in}},\mathcal{V}_{\text{out}}} [ -\sum_{i\in\mathcal{V}_{\text{out}}} \text{CE}(\mY_{i}, \hat{\mY}_{i}) ]\text{ with }\\
\hat{\mY}&=g_{\phi}(\mX \| \mY_{\text{in}}, \mA),
    \end{aligned}
    \label{eq:labeltrick}
\end{equation}
where $\mathcal{V}_{\text{in}}\cup\mathcal{V}_{\text{out}}=\mathcal{V}_{\mathrm{L}}$ are random partitions of $\mathcal{V}_{\mathrm{L}}$, and $\mY_{\text{in}}$ denotes the $N \times C$ matrix, where the rows corresponding to $\mathcal{V}_{\text{in}}$ are taken from $\mY_{\mathrm{L}}$, and the other rows equal $\mathbf{0}$.
Specifying a ratio $\lambda\in(0, 1)$, the random partitions can be sampled by independently drawing from $\text{Bernoulli}(\lambda)$ for each $i\in\mathcal{V}_{\mathrm{L}}$ and including $i$ in $\mathcal{V}_{\text{in}}$ if the outcome is 1.

\noindent\textbf{Reparameterized Discrete Diffusion Models (RDMs).}
In the jargon of discrete diffusion models, we use $\rx^{(t)}$ and $x^{(t)}$ to denote a random variable and its sample at timestep $t$, respectively, where $x^{(t)}$ is a one-hot vector indicating the specific category.
As in common discrete diffusion models~\cite{digress, sink}, the forward (diffusion) process is defined as $q(\rx^{(t)} | \rx^{(t-1)}) = \beta^{(t)}x^{(t-1)} + (1-\beta^{(t)})q_{\text{noise}} $, where $q_{\text{noise}}$ denotes the probabilities of the chosen noise distribution, such as the uniform distribution, and $t\in[T]$ with $[T]$ denotes $\{1,\ldots,T\}$.
This forward transition implies $q(\rx^{(t)} | \rx^{(0)}) = \alpha^{(t)}x^{(0)} + (1-\alpha^{(t)})q_{\text{noise}}$, where $\rx^{(0)}$ obeys the clean data distribution.
The noise schedule $\beta^{(t)}$ and $\alpha^{(t)}=\prod_{s=1}^{t}\beta^{(s)}$ have various design choices, yet they are required to ensure $\alpha^{(t)}$ decrease nearly from 1 to 0 as $t$ increases.

Conventionally, the intractable reverse transition $q(\rx^{(t-1)} | \rx^{(t)})$ is approximated by $q(\rx^{(t-1)} | \rx^{(t)}, \rx^{(0)})$ and therefore estimated as $p_\theta( \rx^{(t-1)} | \rx^{(t)}) := q(\rx^{(t-1)} | \rx^{(t)}, \hat{\rx}^{(0)})$, where a denoiser predicts the clean sample: $\hat{x}^{(0)} = D_{\theta}(x^{(t)}, t)$.
Distinctively, RDMs~\cite{ddm4tg} reformulate the backward (denoising) process, that is, sampling from $q(\rx^{(t-1)} | \rx^{(t)}, \rx^{(0)})$ with a routing mechanism:
\begin{equation}
\begin{aligned}
b^{(t)} &= \1_{x^{(t)}=x^{(0)}},\\
v^{(t-1)} &\sim \text{Bernoulli}(\lambda^{(t-1)}), u^{(t)}\sim q_{\text{noise}},\\
v'^{(t-1)} &\sim \text{Bernoulli}(\lambda'^{(t-1)}), u'^{(t)}\sim q_{\text{noise}}^{x^{(t)}},\\
x^{(t-1)} &= b^{(t)}[v^{(t-1)}x^{(t)} + (1-v^{(t-1)})u^{(t)}]\\
&+ (1-b^{(t)})[v'^{(t-1)}x^{(0)} + (1 - v'^{(t-1)})u'^{(t)} ],
\end{aligned}
    \label{eq:rdmrouting}
\end{equation}
where $q_{\text{noise}}^{x^{(t)}} := \beta^{(t)}x^{(t)} + (1-\beta^{(t)})q_{\text{noise}} $ denotes an interpolation between $x^{(t)}$ and $q_{\text{noise}}$, $\lambda^{(t-1)} := 1 - \frac{ (1-\beta^{(t)})(1-\alpha^{(t-1)})q_{\text{noise}}(\ru=x^{(t)}) }{\alpha^{(t)} + (1-\alpha^{(t)})q_{\text{noise}}(\ru=x^{(t)})} $, and $\lambda'^{(t-1)} := \frac{\alpha^{(t-1)} - \alpha^{(t)}}{1-\alpha^{(t)}}$.
Intuitively, $b^{(t)}$ indicates whether the current sample is noisy; If it is not, $v^{(t-1)}$ determines that it is routed to the current (clean) state or to the noise distribution; otherwise $v'^{(t-1)}$ determines that it is routed to the clean state or to the interpolated noisy distribution.

Furthermore, the routing indicators $\ervv^{(t-1)} := [ v^{(t-1)}, v'^{(t-1)} ]$ can be treated as samples of augmented hidden variable $\rvv^{(t-1)}$.
Then, the reverse transition becomes $q(\rx^{(t-1)}, \rvv^{(t-1)} | \rx^{(t)}, \rx^{(0)})=q(\rvv^{(t-1)})q(\rx^{(t-1)} | \rvv^{(t-1)}, \rx^{(t)}, \rx^{(0)})$, where
\begin{equation}
\begin{aligned}
&q(\rvv^{(t-1)}) =\text{Bernoulli}([\lambda^{(t-1)},\lambda'^{(t-1)}]),\\
&q(\rx^{(t-1)} | \rvv^{(t-1)}, \rx^{(t)}, \rx^{(0)})=\\
&\begin{cases}
    v^{(t-1)}x^{(t)}+(1-v^{(t-1)})q_{\text{noise}} & \text{if }b^{(t)}=1,\\
    v'^{(t-1)}x^{(0)}+(1-v'^{(t-1)})q_{\text{noise}}^{x^{(t)}} & \text{if }b^{(t)}=0.
\end{cases}
\end{aligned}
\label{eq:rdmjoint}
\end{equation}
The demand distribution $q(\rx^{(t-1)} | \rx^{(t)}, \rx^{(0)})$ can be calculated by marginalizing out $\rvv^{(t-1)}$, and then its probabilities would not be changed by the above reparameterization.

Suppose an RDM is applied to model $N$ discrete random variables, we extend our notation to use $\rx_{1:N}^{(t)}$ and $\mathbf{x}_{1:N}^{(t)}$ to denote the variables and their samples, respectively.
Suppose each $\rx_{i}^{(t)}$ is perturbed independently in the diffusion process, with the above reparameterization, the evidence lower bound (ELBO) used to train this RDM can be simplified as follows:
\begin{equation}
\begin{aligned}
&\log{ p(x_{1:N}^{(0)}) } \geq \sum_{t=1}^{T}\mathcal{L}_{t}(\theta)+\text{const., with }\forall t\in[T],\\
&\mathcal{L}_{t}(\theta) = \E_{\prod_{i=1}^{N}q(\rx_{i}^{(t)} | \rx_{1:N}^{(0)})}[ \lambda'^{(t-1)}\sum_{i=1}^{N}(1-b_{i}^{(t)})\text{CE}(x_{i}^{(0)}, \hat{x}_{i}^{(0)}) ],
\end{aligned}
\label{eq:rdmobj}
\end{equation}
where $b_{i}^{(t)}=\1_{x_{i}^{(t)} = x_{i}^{(0)}}$, and $\hat{x}_{1:N}^{(0)} = D_{\theta}(x_{1:N}^{(t)}, t)$.
This form of objective provides a useful property:
\begin{lemma}[Invariance up to reweighting~\cite{ddm4tg}]
    When $q(\rvv^{(t-1)}),t\in[T]$ (see Eq.~\ref{eq:rdmjoint}) considered in the sampling differs from training, Eq.~\ref{eq:rdmobj} is invariant to this difference up to reweighting, i.e., corresponding to different $\lambda'^{(t-1)}$. Moreover, for the multivariate setting, when sampling with $q(\rvv_{i}^{(t-1)}),i\in[N],t\in[T]$ that differs from training, Eq.~\ref{eq:rdmobj} can still be an effective surrogate.
    \label{thm:invariance}
\end{lemma}
As an implication and proof, it has been shown useful to prioritize denoising variables that the denoiser deems more confident in the sampling stage~\cite{ddm4tg}.
\begin{figure*}[t]
    \centering
    \includegraphics[width=0.8\textwidth]{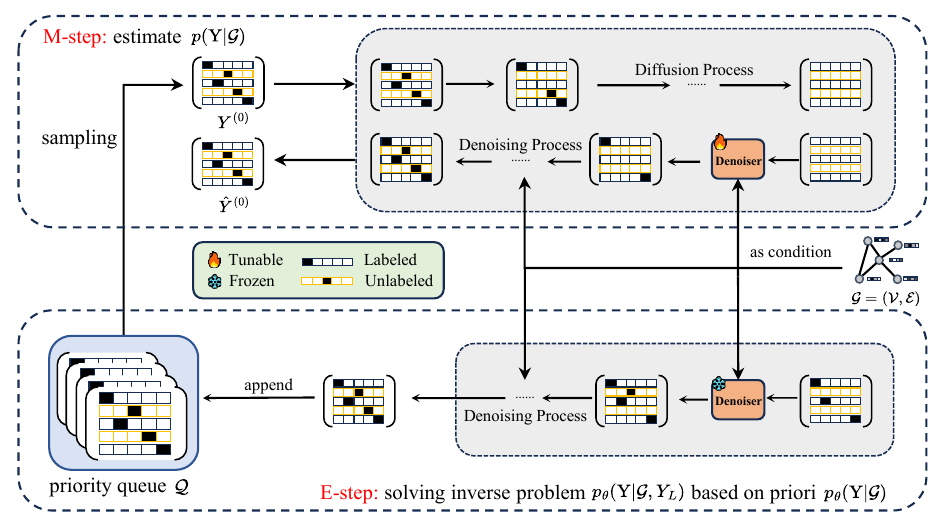}
    \caption{Overview of \ours: a variational EM framework that alternates between pseudo-label inference (E-step) and diffusion model training (M-step).}
    \label{fig:framework}
\end{figure*}

\section{Methodology}
\label{sec:method}
To enable structured prediction for scalable node classification, we present \ours, which employs a specific instance of RDM, namely reparameterized masked diffusion model (RMDM) to estimate $p(\rmY|\gG)$, the joint probability distribution of all node labels given the graph.
However, only $\mY_{\mathrm{L}}$ is observed during training, while $\mY_{\mathrm{U}}$ remains unobserved, making it infeasible to estimate $p(\rmY|\gG)=p(\rmY_{\mathrm{L}}, \rmY_{\mathrm{U}}|\gG)$ directly.
To address this, we adopt the variational expectation-maximization (EM) framework, which has been shown to be effective in similar scenarios~\cite{VEM,DPM-SNC}.

Sec.~\ref{subsec:overview} outlines our overall formulation under the variational EM framework.
The M-step and E-step are detailed in Sec.~\ref{subsec:M-step} and Sec.~\ref{subsec:E-step}, respectively.
Sec.~\ref{subsec:analysis} provides a theoretical analysis of \ours, highlighting its advantages and connections to related work.

\subsection{Overview}
\label{subsec:overview}
We present an overview of \ours in Fig.~\ref{fig:framework}.
We treat the unlabeled node labels $\rmY_{\mathrm{U}}$ as hidden variables and model the joint distribution $p(\rmY_{\mathrm{L}}, \rmY_{\mathrm{U}}|\gG)$, or equivalently, $p(\rmY | \gG)$ using a RMDM parameterized by $\theta$.
Direct maximum likelihood estimation of $p_{\theta}(\rmY_{\mathrm{L}} | \gG)$ involves marginalizing over all possible label assignments for hidden variables, making it computationally infeasible.
Thus, we adopt a variational expectation-maximization (EM) framework, where we iteratively alternate between the E-step and the M-step to maximize the evidence lower bound (ELBO):
\begin{equation}
\begin{aligned}
    \mathcal{L}_{\mathrm{MLE}}(\theta)=&\mathcal{L}_{\text{ELBO}}(\theta,q) + \KL(q(\rmY_{\mathrm{U}})\|p_{\theta}(\rmY_{\mathrm{U}}|\gG,\mY_{\mathrm{L}})) \\
    =&\E_{\mY_{\mathrm{U}}\sim q(\rmY_{\mathrm{U}})}[ \log{p_{\theta}(\rmY_{\mathrm{L}}=\mY_{\mathrm{L}}, \rmY_{\mathrm{U}}=\mY_{\mathrm{U}} | \gG)}]\\
    &-\E_{\mY_{\mathrm{U}}\sim q(\rmY_{\mathrm{U}})}[ \log{ p_{\theta}(\rmY_{\mathrm{U}} = \mY_{\mathrm{U}} | \gG, \mY_{\mathrm{L}}) } ] ,
\label{eq:vlb}
\end{aligned}
\end{equation}
where $q(\cdot)$ stands for the variational distribution~\footnote{In conditional probability expressions such as $p_{\theta}(\rmY_{\mathrm{U}} | \gG, \mY_{\mathrm{L}})$, we write the conditioning variables directly in terms of their realizations, omitting explicit equations like $\rmY_{\mathrm{U}}=\mY_{\mathrm{U}}$, to keep our notion concise. This convention is used consistently throughout the paper.}.

\noindent\textbf{E-step (Pseudo-Label Inference)}: 
We aim to sample pseudo-labels for unlabeled nodes based on current optimal posterior $p_{\theta}(\rmY_{\mathrm{U}} | \gG, \mY_{\mathrm{L}})$.
To this end, we freeze the current RMDM and sample $\tilde{\mY}\sim p_{\theta}(\rmY |\gG, \mY_{\mathrm{L}})$, which corresponds to solving an inverse problem (i.e., label completion), enforcing consistency with known labels.
This procedure tightens the ELBO by ensuring the $q(\rmY_{\mathrm{U}})$ aligns closely with the true posterior.

\noindent\textbf{M-step (Diffusion Model Training)}:
We aim to maximize the ELBO w.r.t. $\theta$ based on current $q(\rmY_{\mathrm{U}})$.
To this end, we treat pseudo-label samples from the E-step as observed training data and update $\theta$ by maximizing their likelihood under our RMDM. This directly increases the ELBO.

Intuitively, performing these two steps alternatively facilitates mutual refinement:
better pseudo-labels improve model training, and a better-trained model generates more accurate pseudo-labels. The overall learning procedures are summarized in Alg.~\ref{alg:vem}. At test time, only the E-step is performed to generate predictions.

\begin{algorithm}[t]
\caption{Learning procedure for \ours.}
\label{alg:vem}
\newcommand{\INDSTATE}[1][1]{\STATE\hspace{#1\algorithmicindent}}
\renewcommand\algorithmicrequire{\textbf{Input:}}
\begin{algorithmic}[1]
\REQUIRE Graph $\gG$, observed labels $\mY_{\mathrm{L}}$
\STATE \textbf{Initialize priority queue:}
\INDSTATE $\mathcal{Q}=\emptyset$
\INDSTATE Train a GNN $g_{\phi}(\mX,\mA)$ based on Eq.~\ref{eq:gnnobj}
\INDSTATE \textbf{for} $i = 1$ to $S$ \textbf{do}
\INDSTATE[2] Sample $\tilde{\mY}$ from $p_{\phi}(\rmY|\gG)$ based on Eq.~\ref{eq:independence}
\INDSTATE[2] Enforce constraint: $\tilde{\mY}_{\mathrm{L}} \gets \mY_{\mathrm{L}}$
\INDSTATE[2] Push $\tilde{\mY}$ into $\mathcal{Q}$
\INDSTATE \textbf{end}

\STATE Initialize RMDM's parameter $\theta$

\REPEAT
    \STATE \textbf{E-step: Pseudo-Label Inference}
    \INDSTATE $\tilde{\mY} \gets$ \textsc{Sample}$(\gG, \mY_{\mathrm{L}}, \theta)$
    \INDSTATE Push $\tilde{\mY}$ into $\mathcal{Q}$
    \STATE \textbf{M-step: Diffusion Model Training}
    \INDSTATE Sample $\mY^{(0)}$ from $\mathcal{Q}$
    \INDSTATE Compute $\mathcal{L}(\theta)$ with $\mY^{(0)}$ based on Eq.~\ref{eq:denoise_los}
    \INDSTATE Update $\theta$ based on $\nabla_{\theta}\mathcal{L}(\theta)$
\UNTIL \textit{stopping criterion is met}
\RETURN learned RMDM $\theta$
\end{algorithmic}
\end{algorithm}

\begin{algorithm}[t]
\caption{\textsc{Sample}($\gG, \mY_{\mathrm{L}}, \theta$)}
\label{alg:sample}
\begin{algorithmic}[1]
\REQUIRE Graph $\gG$, observed labels $\mY_{\mathrm{L}}$, model parameter $\theta$
\STATE Initialize $\mY^{(T)}$ with $\mY_{i}^{(T)} \gets \mathbf{1}_{\text{sink}},i\in[N]$
\FOR{timestep $t = T$ to $1$}
    \STATE Sample $\ervv_{i}^{(t-1)} \sim q(\rvv_{i}^{(t-1)}),i\in[N]$ based on Eq.~\ref{eq:denoisingeachcond}
    \STATE Apply labeled-first strategy to modify $\ervv_{i}^{(t-1)},i\in[N]$
    \STATE $\hat{\mY}^{(0)} \gets D_{\theta}(\mY^{(t)}, t, \gG)$
    \STATE Sample $\mY^{(t-1)} \sim q(\rmY^{(t-1)} | \rvv_{1:N}^{(t-1)}, \mY^{(t)}, \hat{\mY}^{(0)})$ based on Eq.~\ref{eq:denoisingeachcond} and Eq.~\ref{eq:denoisingcond}
\ENDFOR
\RETURN $\mY^{(0)}$
\end{algorithmic}
\end{algorithm}

\subsection{M-step: Diffusion Model Training}
\label{subsec:M-step}

Suppose that we have acquired a sample $\tilde{\mY}$ that takes the same values as $\mY_{\mathrm{L}}$ on $\mathcal{V}_{\mathrm{L}}$ while its values on $\mathcal{V}_{\mathrm{U}}$ can be regarded as an unbiased sample of current $q(\rmY_{\mathrm{U}})$.
Now we treat this sample as drawn from the true data distribution and update our RMDM to better estimate it.

\subsubsection{Diffusion and Denoising Processes}
\label{subsubsec:processes}
For a RDM (see Sec.~\ref{sec:pre}), there are two popular design choices of $q_{\text{noise}}$: a uniform distribution~\cite{uniform_distribution} or a point mass on the special ``sink'' state~\cite{sink}.
We adopt the latter, and thus the outcome space is augmented with a ``sink'' state represented as follows:
\begin{equation}
\mathbf{1}_{\text{sink}}:=
\begin{pmatrix}
\overbrace{0, 0, \ldots, 0}^{C \text{ zeros}}, 1
\end{pmatrix}.
\label{eq:sinkstate}
\end{equation}
Then, each $\mY_i$ becomes a $(C+1)$-dimensional one-hot vector.

The diffusion process progressively masks the node labels until all the node labels are completely masked, where once a node is masked, it transits to and stays in $\mathbf{1}_{\text{sink}}$.
Thus, we refer to our model the reparameterized masked diffusion model (RMDM).
Specifically, we choose to diffuse node labels independently, that is, $q(\rmY^{(t)} | \rmY^{(t-1)}) := \prod_{i\in\mathcal{V}} q(\rmY_{i}^{(t)} | \rmY^{(t-1)}),t\in [T]$, with $q(\rmY_{i}^{(t)} | \rmY^{(t-1)}) := \beta^{(t)} \mY_{i}^{(t-1)} + (1-\beta^{(t)})\mathbf{1}_{\text{sink}}$ for any $i\in\mathcal{V}$ and timestep $t\in [T]$.
As a result, we can independently sample for each $\rmY_{i}^{(t)}$ based on the probability distribution as follows:
\begin{equation}
q(\rmY_{i}^{(t)} | \rmY^{(0)} ) = \alpha^{(t)} \mY_{i}^{(0)} + (1 - \alpha^{(t)}) \mathbf{1}_{\text{sink}},
\label{eq:DDM_forward}
\end{equation}
where $\alpha^{(t)}$ controls the expected proportion of nodes that remain intact at timestep $t$.
Here we adopt the cosine schedule~\cite{cosine},
which ensures that $\alpha^{(t)}$ decreases from 1 to 0 as $t$ increases, satisfying the condition stated in Sec.~\ref{sec:pre}.
Consequently, as $t$ increases, there are more and more node labels masked in $\mY^{(t)}\sim q(\rmY^{(t)} | \rmY^{(0)} )=\prod_{i\in\mathcal{V}}q(\rmY_{i}^{(t)} | \rmY^{(0)})$ until all are $\mathbf{1}_{\text{sink}}$.

To perform the denoising process, the denoiser is expected to predict the clean sample: $\hat{\mY}^{(0)} = D_{\theta}(\mY^{(t)}, t, \gG)$, where $\mY^{(t)}$ is the current sample drawn at the timestep $t$ and $\gG$ serves as a condition.
In particular, we restrict the output of the denoiser to take values from $\mathcal{Y}$ without $\mathbf{1}_{\text{sink}}$ and force $\forall i\in\mathcal{V}, \hat{\mY}_{i}^{(0)}=\mY_{i}^{(t)}$ if $\mY_{i}^{(t)}\neq \mathbf{1}_{\text{sink}}$.
The reason is that the clean sample does not take value from the ``sink'' state, and our diffusion process masks each node's label once and only once so that the denoising step should also take place only once.

Moreover, consider Eq.~\ref{eq:rdmrouting} but extend it to the multivariate case, $b_{i}^{(t)}=1$ implies that $\mY_{i}^{(t)}\neq \mathbf{1}_{\text{sink}}$, that is, the $i$-th node's label has been denoised.
In this case, $q_{\text{noise}}(\ru=\mY_{i}^{(t)})=0$, and thus $\lambda_{i}^{(t-1)}=1$, which ensures that the $i$-th node's label would not be re-pertubed.
Thus, in our RMDM, Eq.~\ref{eq:rdmjoint} can be written as follows:
\begin{equation}
\begin{aligned}
&q(\rvv_{i}^{(t-1)})=q([\rv_{i}^{(t-1)}, \rv_{i}^{\prime (t-1)}]) = \text{Bernoulli}([1, \lambda_{i}^{\prime (t-1)}]),\\
&q(\rmY_{i}^{(t-1)} | \rvv_{i}^{(t-1)}, \rmY_{i}^{(t)}, \rmY_{i}^{(0)} )=\\
&\begin{cases}
\mY_{i}^{(t)},&\text{if } b_{i}^{(t)}=1,\\[10pt]
\begin{aligned}
& v_{i}^{\prime (t-1)} \mY_{i}^{(0)} \\
&+ (1-v_{i}^{\prime(t-1)})
\bigl(\beta_{i}^{(t)} \mY_{i}^{(t)} + (1-\beta_{i}^{(t)})\mathbf{1}_{\text{sink}} \bigr)
\end{aligned},&\text{if }b_{i}^{(t)}=0.
\end{cases}
\end{aligned}
\label{eq:denoisingeachcond}
\end{equation}
Here we use subscript $i$ to emphasize that node-wise noise schedule is allowed.
The reverse transition can be decomposed as follows:
\begin{equation}
\begin{aligned}
&q(\rmY^{(t-1)}, \rvv_{1:N}^{(t-1)} | \rmY^{(t-1)}, \rmY^{(0)})\\
=&q(\rvv_{1:N}^{(t-1)})q(\rmY^{(t-1)} | \rvv_{1:N}^{(t-1)}, \rmY^{(t-1)}, \rmY^{(0)})\\
=&\prod_{i\in\mathcal{V}}q(\rvv_{i}^{(t-1)})q(\rmY_{i}^{(t-1)} | \rvv_{i}^{(t-1)}, \rmY_{i}^{(t)}, \rmY_{i}^{(0)} ).
\end{aligned}
\label{eq:denoisingcond}
\end{equation}
Consequently, the denoising process starts with sampling $\mY^{(T)}$ from $p(\rmY^{(T)})=\prod_{i\in\mathcal{V}}p(\rmY_{i}^{(T)}) = \prod_{i\in\mathcal{V}}\delta(\mathbf{1}_{\text{sink}})$, where $\delta(\cdot)$ denotes the Dirac delta distribution.
Then we iteratively sample $\mY^{(t-1)}$ according to Eq.~\ref{eq:denoisingeachcond} and Eq.~\ref{eq:denoisingcond} until we get a fully denoised sample at $t=0$.

\subsubsection{Time-aware GNN Layers for the Denoiser}
\label{subsec:na}
\begin{figure}
    \centering
    \includegraphics[width=0.48\textwidth]{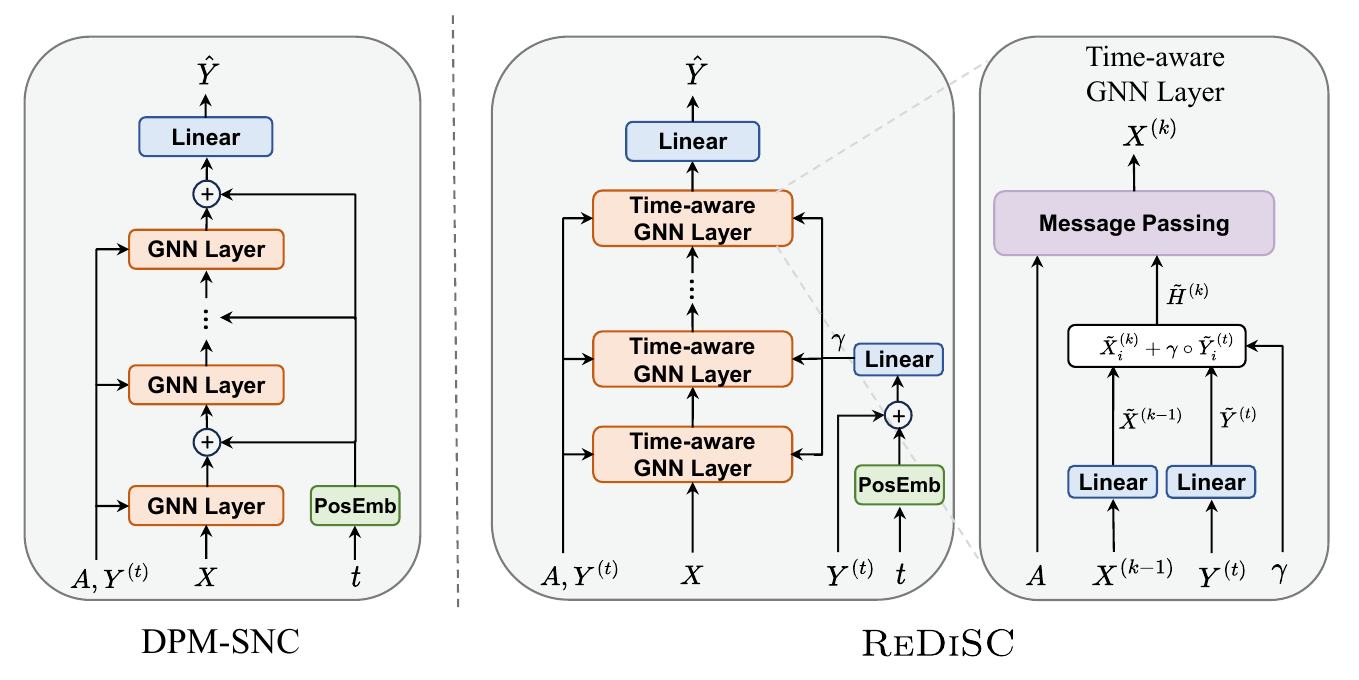}
    \caption{Comparison of the neural architectures: \ours fuses features and labels via time-aware scaling.}
    \label{fig:architecture}
\end{figure}

Existing diffusion models often incorporate the timestep $t$ as an input to the denoiser $D_{\theta}(\cdot)$ so that it can refine the noisy sample in a time-aware manner.
Both DPM-SNC~\cite{DPM-SNC} and our proposed \ours follow this convention, and we compare their neural architectures in Fig.~\ref{fig:architecture}.

DPM-SNC stacks vanilla GNN layers to form the denoiser while transforming the sinusoidal positional encoding of timestep and then adding this transformed representation to the output node embeddings of each GNN layer.
However, this temporal information has not been utilized in forward propagation within each GNN layer, which could lead to insufficient interaction among node features, noisy node labels, and timestep.

In contrast, \ours makes each GNN layer time-aware by using the concatenation of transformed sinusoidal positional encoding of timestep $t$ and noisy labels $\mY^{(t)}$ to produce a scaling factor $\gamma\in[0, 1]^{d'}$, where $d'$ denotes the dimension of the node/label embedding. Then suppose that the linearly transformed node and label embeddings of the $i$-th node in the $k$-th GNN layer is $\tilde{\mX}_{i}^{(k-1)}$ and $\tilde{\mY}_{i}^{(t)}$, this scaling factor modulates the fusion between them, producing an adjusted representation $\tilde{\mH}_{i}^{(k)} = \tilde{\mX}_{i}^{(k-1)} + \gamma \circ \tilde{\mY}_{i}^{(t)}$, where $\circ$ denotes elementwise multiplication. The updated features $\tilde{\mH}^{(k)}$ as well as the adjacency matrix $\mA$ are then used in the forward propagation of the $k$-th GNN layer to produce ${\mX}^{(k)}$.

\subsubsection{Optimization}
\label{subsubsec:opt}
To maximize $\E_{\tilde{\mY}\sim \delta(\mY_{\mathrm{L}})\times q(\rmY_{\mathrm{U}})}[ \log{ p_{\theta}(\rmY=\tilde{\mY} | \gG ) } ]$, we treat $\tilde{\mY}$ as an unbiased sample of the true data distribution our RMDM aims to estimate.
Then, we employ Eq.~\ref{eq:rdmobj} and update our RMDM by minimizing the objective as follows:
\begin{equation}
\begin{aligned}
\mathcal{L}(\theta) &= \E_{t \sim \mathcal{U}([T])} [ \mathcal{L}_{t}(\theta) ]\text{ with }\forall t\in[T],\\
\mathcal{L}_{t}(\theta)
&= \E_{\prod_{i\in\mathcal{V}} q(\rmY_{i}^{(t)} \mid \rmY^{(0)}=\tilde{\mY})}
\\
&\quad \bigg[
    -\sum_{i\in\mathcal{V}}
    \lambda_{i}^{\prime(t-1)} (1 - b_{i}^{(t)})
    \text{CE}(\tilde{\mY}_{i}, \hat{\mY}_{i}^{(0)})
\bigg],
\end{aligned}
\label{eq:denoise_los}
\end{equation}
where $\mathcal{U}([T])$ denotes a uniform distribution over $[T]$, and $\hat{\mY}^{(0)}=D_{\theta}(\mY^{(t)}, t, \gG)$ is our denoiser's predicted clean sample.
Since sampling from $ q(\rmY^{(t)} | \rmY^{(0)}=\tilde{\mY})$ can be performed according to Eq.~\ref{eq:DDM_forward}, we can efficiently estimate $\nabla_{\theta}\mathcal{L}(\theta)$ and apply Adam optimizer~\cite{adam} to update our denoiser.


\subsection{E-step: Pseudo-Label Inference}
\label{subsec:E-step}
As we have summarized in Sec.~\ref{subsec:overview}, our E-step aims to fill the gap between $\mathcal{L}_{\text{ELBO}}(\theta, q)$ and $\mathcal{L}_{\text{MLE}}(\theta)$ by minimizing the last term in Eq.~\ref{eq:vlb} w.r.t. $q(\cdot)$.
Then we need to sample from $q(\cdot)$.
Thus, we choose to draw sample from the minimizer, i.e., $q(\rmY_{\mathrm{U}}) = p_{\theta}( \rmY_{\mathrm{U}} | \gG, \mY_{L} )$.

\subsubsection{Labeled-first Inference for Inverse Problem Solving}
\label{subsubsec:invprob}

At first, sampling from $p_{\theta}( \rmY_{\mathrm{U}} | \gG, \mY_{\mathrm{L}} )$ is equivalent to drawing sample $\tilde{\mY}$ from $p_{\theta}( \rmY | \gG, \mY_{L} )$ and taking $\tilde{\mY}_{\mathrm{U}}$, which obeys the desired probability distribution $p_{\theta}( \rmY_{\mathrm{U}} | \gG, \mY_{L} )$.
Thus, we turn to sampling from $p_{\theta}( \rmY | \gG, \mY_{L} )$, which means solving the label completion problem, i.e., an inverse problem for the given priori $p_{\theta}( \rmY | \gG)$.


At first, Eq.~\ref{eq:DDM_forward} implies that, during the denoising process, the state of a random variable should only change once---from $\mathbf{1}_{\text{sink}}$ to one of $\mathcal{Y}$ and remain unchanged.
In other words, for any $i\in\mathcal{V}_{\mathrm{L}}$, if $\mY_{i}^{(t)} \neq \mathbf{1}_{\text{sink}} \land \mY_{i}^{(t)}\neq \mY_i$, this stochastic path would not lead to any valid sample.
Hence, at each denoising transition, we draw $\ervv_{i}^{(t-1)}=[v_{i}^{(t-1)}, v_{i}^{\prime (t-1)}]$ according to Eq.~\ref{eq:denoisingeachcond} for each node $i$ that satisfies $\mY_{i}^{(t)}=\mathbf{1}_{\text{sink}}$.
For nodes that have $v_{i}^{\prime (t-1)}=1$, we sample $\mY_{i}^{(t-1)}$ from $\hat{\mY}_{i}^{(0)}$ and keep the state of other nodes unchanged.
To sample from $p_{\theta}(\rmY|\gG,\mY_{\mathrm{L}})$, we can repeatedly simulate the above process to sample from $p_{\theta}(\rmY | \gG)$, while pruning the stochastic path once we encounter $\mY_{i}^{(t-1)} \neq \mY_{i}$ for any $i\in\mathcal{V}_{\mathrm{L}}$.

However, the pruning might occur at small timesteps, which will waste a large fraction of computations in performing the simulation.
To improve sampling efficiency, we introduce the labeled-first inference strategy.
Specifically, we keep $\sum_{i:\mY_{i}^{(t)}=\mathbf{1}_{\text{sink}}}\lambda_{i}^{\prime (t-1)}$, the expected number of to be denoised nodes, unchanged yet allocate this denoising budget to $\mathcal{V}_{\mathrm{L}}$ in priority until all $\mY_{\mathrm{L}}^{(t)}$ have been denoised.
Hence, we always perform pruning at the early stage of denoising process, which enables that $\forall i\in\mathcal{V}_{\mathrm{L}}$, if $\mY_{i}^{(t)} = \mathbf{1}_{\text{sink}} \wedge v_{i}^{\prime (t-1)}=1$, we set $\mY_{i}^{(t-1)} = \mY_{i}$ whatever $\hat{\mY}_{i}^{(0)}$ is.
In each E-step, we perform denoising process with this labeled-first inference to sample one $\tilde{\mY}$ from $p_{\theta}(\rmY | \gG, \mY_{\mathrm{L}})$ as summarized in Alg.~\ref{alg:sample}.

As $\theta$ is just updated in the last M-step, this sampled $\tilde{\mY}$ tends to be different from that of the previous E-step to some extent.
As these samples are used to update $\theta$ in the M-step, this discrepancy in consecutive E-steps should be controlled to avoid unstable training.


\subsubsection{Priority Queue for Sample Selection}
\label{subsubsec:priority_queue}

To improve stability, we follow DPM-SNC~\cite{DPM-SNC} to cache the samples drawn in recent $S$ E-steps in a queue $\mathcal{Q}$ and then, in each M-step, randomly select one $\tilde{\mY}$ from $\mathcal{Q}$.
Obviously, a larger queue size $S$ results in a more stable data distribution, and a smaller $S$ corresponds to a less stable but more up-to-date data distribution.
Instead of uniformly selecting one $\tilde{\mY}$ from $\mathcal{Q}$, we argue that samples corresponding to a more accurate prediction should have a greater chance of being selected because, in the M-step, the RMDM is updated to approximate the probability distribution of these samples.

Inspired by prioritized experience replay mechanism~\cite{prioritized}, we propose to cache samples in a priority queue $\mathcal{Q}$.
Specifically, when a sample $\tilde{\mY}$ has been drawn, we evaluate its accuracy on the validation set and assign this validation accuracy to it as its priority.
Then we add this $\tilde{\mY}$ to $\mathcal{Q}$, where the most dated sample will be excluded if there have been $S$ elements in $\mathcal{Q}$.
In the M-step, we randomly pick one $\tilde{\mY}$ from $\mathcal{Q}$, where each cached sample's likelihood is proportional to its priority with a temperature $\tau$.
As usual, a larger $\tau$ makes this distribution more uniform, and a smaller $\tau$ means more reliance on these priorities.




\subsection{Theoretical Analysis}
\label{subsec:analysis}
In this section, we comprehensively analyze \ours from the following aspects.

\noindent\textbf{Structured Prediction.}
We train our RMDM with $\mathcal{L}(\theta)$ defined in Eq.~\ref{eq:denoise_los}, which is also a sum of cross-entropy losses as $\mathcal{L}_{\text{GNN}}(\phi)$ defined in Eq.~\ref{eq:gnnobj}.
Although this form of objective reflects the independence assumption among $p_{\theta}(\rmY_{i}^{(t-1)} | \rmY^{(t)}),i\in[N]$, our RMDM can still express the dependencies between $\rmY^{(0)}$'s components.
\begin{proposition}
When $p(\rmY^{(T)})=\prod_{i\in\mathcal{V}}p(\rmY_{i}^{(T)})$ and $p_{\theta}(\rmY^{(t-1)} | \rmY^{(t)},\gG)=\prod_{i\in\mathcal{V}}p_{\theta}(\rmY_{i}^{(t-1)} | \rmY^{(t)}, \gG)$, components of $p_{\theta}(\rmY^{(0)} | \gG)$ can still be correlated.
    \label{thm:}
\end{proposition}
The proof is straightforward.
Diffusion models express the data distribution as 
$p_{\theta}(\rmY^{(0)} | \gG) = \sum_{\mY^{(1:T)}}p_{\theta}(\rmY^{(0)} | \mY^{(1)}, \gG)p(\rmY^{(T)})\prod_{t=2}^{T}p_{\theta}(\rmY^{(t-1)}| \mY^{(t)}, \gG)$.
In vanilla GNNs, both $\rmY_{i}$ and $\rmY_{j}$ are conditioned on $\gG$, forming d-separation.
In diffusion models, $\rmY_{i}^{(t-1)}$ and $\rmY_{j}^{(t-1)}$ are additionally conditioned on $\rmY^{(t)}$ that will be marginalized out, turning $\rmY_{i}^{(t-1)}$ and $\rmY_{j}^{(t-1)}$ d-connected and thus non-independent~\cite{pgm}.
Our RMDM also expresses $p_{\theta}(\rmY | \gG)$ in this way, enabling \ours to produce structured prediction as DPM-SNC~\cite{DPM-SNC} that considers diffusion models in the continuous domain.

\noindent\textbf{Inverse Problem Solving.}
In our E-step, we modify $q(\rv_{i}^{\prime (t-1)})$s so that we can efficiently solve the inverse problem with Alg.~\ref{alg:sample}.
As presented in Sec.~\ref{subsubsec:invprob}, we keep the expected number of to be denoised nodes unchanged at current timestep $t$.
Accordingly, the corresponding mask ratio of $\mY^{(t)}$ in diffusion process is unchanged.
The only change lies in $q(\rv_{i}^{\prime (t-1)})$s, that is, masking $\mathcal{V}_{\mathrm{U}}$ in priority.
Once $D_{\theta}(\cdot)$ can generalize well with this distribution shift, $\mathcal{L}(\theta)$ in Eq.~\ref{eq:denoise_los} is obviously an effective surrogate based on Lemma~\ref{thm:invariance}.
Although this assumption looks unrealistic, both RDM's certainty-first inference strategy~\cite{ddm4tg} and our RMDM's labeled-first inference strategy perform well in practice.

\noindent\textbf{Time Complexity.}
The property of RMDM that each node will be denoised once and only once during the denoising process (see Sec.~\ref{subsubsec:invprob}) enables scalable implementation for \ours.
\begin{proposition}
    Suppose the average degree of $\gG$ is $D$, and the denoiser is a $K$-layer GNN, then the E-step of DPM-SNC needs $O(TNDK)$ message-passing operations, while that of \ours needs $O(\min(TNDK, ND^{K}))$.
    \label{thm:timecplx}
\end{proposition}
It is well known that the forward propagation of a vanilla GNN layer on all nodes involves $|\mathcal{E}|=ND$ message-passing operations. Then, calculating $\hat{\mY}^{(0)}=D_{\theta}(\mY^{(t)}, t, \gG)$ needs $NDK$ such operations.
Since DPM-SNC, in each timestep, calculates not only $\hat{\mY}^{(0)}$ but also $\nabla_{\mY^{(t)}}\| \hat{\mY}_{\mathrm{L}}^{(0)} - \mY_{\mathrm{L}} \|_{2}^{2}$, its complexity is $O(TNDK)$ with a large constant due to backward propagation.
In contrast, \ours only needs to make inference for each node once during the denoising process.
Therefore, it can be implemented as iteratively making inference for each node with graph sampling~\cite{GraphSAGE}, which corresponds to a complexity $O(ND^K)$.
This result looks worse, but, on practical large graphs~\cite{OGB}, $D$ is often not large (e.g., $D=5$) and $K=2$ is usually deep enough.
In this case, $D^K < TDK$ because, in general, the horizon $T$ cannot be quite small (e.g., $T=80$ in DPM-SNC).
Besides, in each timestep, the neighborhoods of to be denoised nodes tend to have overlap, which makes $ND^K$ significantly overestimate the practical complexity.

\noindent\textbf{Objective.}
In \ours's implementation, we represent the masked node label as a $c$-dimensional zero vector rather than that in Eq.~\ref{eq:sinkstate}, which makes $\mX\|\mY^{(t)}$, the input to our GNN denoiser $D_{\theta}(\cdot)$, in the same form as that of $g_{\phi}(\cdot)$ in Eq.~\ref{eq:labeltrick}.
Besides, both their objectives are weighted sums of cross-entropy losses, which leads to the connection as follows:
\begin{proposition}
Assume that $D_{\theta}(\cdot)$ is sufficiently expressive to function as if it had separate parameters $\theta^{(t)}$ for each timestep $t$~\cite{faster}, our learned $p_{\theta}(\rmY|\gG)$ can be interpreted as an ensemble of the ``GNN+LP'' hybrids (see Sec.~\ref{sec:pre}) that correspond to different mask ratios.
\end{proposition}
Obviously, in Eq.~\ref{eq:denoise_los}, $\mathcal{L}_{t}(\theta)$ can be re-written as:
\begin{equation}
\begin{aligned}
&\mathcal{L}(\theta) = \E_{t \sim \mathcal{U}([T])} [ \mathcal{L}_{t}(\theta) ]\\
=&\E_{t \sim \mathcal{U}([T])} [ -\lambda^{\prime (t)}\E_{\mathcal{V}_{\text{in}}, \mathcal{V}_{\text{out}} }[ \sum_{i\in\mathcal{V}_{out}}\text{CE}(\tilde{\mY}_{i},\hat{\mY}_{i}^{(0)})] ]\\
=&\E_{t \sim \mathcal{U}([T])} [ -\lambda^{\prime (t)}\mathcal{L}_{\text{GNN+LP}}^{\alpha^{(t)}}(\theta) ],
\end{aligned}
    \label{eq:connection}
\end{equation}
where $\mathcal{V}_{\text{in}}\cup\mathcal{V}_{\text{out}}$ are random partitions of $\mathcal{V}$ with $\mathcal{V}_{\text{out}}=\{i\in\mathcal{V} : b_{i}^{(t)}=1\}$, that is, $\mY_{i}^{(t)} = \tilde{\mY}_{i}, \mY^{(t)}\sim q(\rmY^{(t)} | \rmY^{(0)}=\tilde{\mY})$.
With that conventional assumption, \ours ensembles $T$ ``GNN+LP'' hybrids.
\section{Related Work}
\label{sec:related}

\noindent \textbf{Graph Neural Networks.}
Graph Neural Networks (GNNs) have become the dominant paradigm for node classification by effectively integrating graph topology and node attributes through neighborhood aggregation. Representative models, including GCN~\cite{GCN}, GraphSAGE~\cite{GraphSAGE}, GAT~\cite{GAT}, and GIN~\cite{GIN}, typically update node representations via $k$-hop message passing to facilitate node classification. However, these methods generally assume conditional independence of node labels given the learned representations, ignoring structured dependencies. Moreover, shallow architectures and oversmoothing issues~\cite{oversmoothing} limit their ability to capture long-range interactions. Although strategies such as label reuse~\cite{LabelReuse}, combining GNNs with LP regularization~\cite{LPA}, deeper GNNs with residual connections~\cite{JKNet}, and graph transformers~\cite{Graphormer} have been explored, explicit structured label reasoning remains underdeveloped.

\noindent \textbf{Label Propagation Methods.}
Label propagation (LP) infers labels by diffusing information from labeled to unlabeled nodes, leveraging the homophily assumption. While efficient and interpretable, LP does not incorporate node features or adapt effectively to complex graph structures. Extensions such as TLLT~\cite{TLLT} adapt propagation dynamics based on sample reliability to mitigate noise, while LD~\cite{label_deconvolution} and the label trick~\cite{parallel_propagate} reinterpret LP within frameworks that blend structural consistency with feature learning. In transductive contexts, LP often provides pseudo-labels for data augmentation, as in PTA~\cite{PTA}, improving performance under limited annotations.

\noindent \textbf{Structured Prediction with GNNs.}
Structured prediction methods enhance node classification by explicitly modeling label dependencies, typically through joint label distributions. Early approaches like GMNN~\cite{GMNN} combine GNNs with CRFs in a variational EM framework to leverage both local features and global label interactions. SPN~\cite{SPN} advances this idea by learning CRF potentials with neural networks, enabling more flexible yet tractable inference. G3NN~\cite{G3NN} further extends to jointly modeling node attributes, labels, and structure under a unified generative framework. Alternatively, CL~\cite{CL} adopts Monte Carlo sampling to capture label dependencies in inductive settings. Despite their differences in formulation, these methods commonly rely on sophisticated inference schemes, which can hinder scalability and practical deployment.

\noindent \textbf{Diffusion-Based Models for Node Classification.}
Diffusion probabilistic models (DPMs)~\cite{DDPM} have recently emerged as a powerful generative framework for learning complex data distributions, including graphs.
In node classification tasks, DPM-SNC~\cite{DPM-SNC} applies this paradigm to learn the joint distribution of node labels through a discrete-time diffusion process, incorporating manifold constraints and label conditioning for structured predictions. LGD~\cite{LGD} extends this approach with specialized encoder-decoder architectures and cross-attention to better capture structural patterns and unify various graph learning tasks.
Despite their expressiveness, diffusion models are often computationally intensive, struggle to align continuous representations with discrete labels, and lack interpretability or clear ties to classical LP and GNN methods. These issues motivate the development of discrete diffusion approaches that preserve generative strengths while improving efficiency and structural alignment.

\section{Experiments}
\label{sec:exp}

We conduct comprehensive experiments to evaluate the effectiveness of \ours on a wide range of transductive node classification tasks, spanning both homophilic and heterophilic graphs of varying sizes.
We also perform ablation studies to isolate the contributions of key design choices in \ours, including the priority queue, the labeled-first inference, and the time-aware GNN layer.
Additionally, we analyze the model's sensitivity to the number of diffusion steps to assess its robustness under coarser discretization.
All experiments are conducted on a machine equipped with eight NVIDIA TESLA V100 GPUs (32 GB), Intel Xeon Platinum 8163
CPU (2.50 GHz), and 724 GB RAM.

\begin{table*}[t]
\centering
\caption{Accuracies on homophilic graphs at both node-level (N-Acc.) and subgraph-level (Sub-Acc.).}
\label{tab:homo_results}
\resizebox{\textwidth}{!}{
\begin{tabular}{@{}ccccccccccc@{}}
\toprule
         & \multicolumn{2}{c}{PubMed} & \multicolumn{2}{c}{Cora} & \multicolumn{2}{c}{Citeseer} & \multicolumn{2}{c}{Photo} & \multicolumn{2}{c}{Computer} \\ 
Method   & N-Acc.       & Sub-Acc.    & N-Acc.      & Sub-Acc.   & N-Acc.        & Sub-Acc.     & N-Acc.      & Sub-Acc.    & N-Acc.        & Sub-Acc.     \\ \midrule
LP       & 71.60±0.0            & 49.90±0.0           & 71.00±0.0           & 51.10±0.0          & 49.80±0.0             & 32.70±0.0            & 81.76±3.49           & 42.93±3.08           & 72.07±2.60             & 19.67±1.64            \\
PTA      & 78.08±0.28            & 55.24±0.13           & 82.05±0.28           & 62.06±0.35          & 72.33±0.61             & 51.94±0.58            & 90.11±1.52           & 51.04±1.46           & 81.62±1.67             & 26.29±1.02            \\ 
\midrule
LGD   & 70.74±1.36           & 45.18±1.53           & 71.26±2.13           & 53.45±1.22          & 56.30±1.53             & 35.67±0.96          & 90.39±0.97           & 53.14±0.81           & 81.28±1.96             & 26.63±1.19            \\

\midrule
GCN      & 78.96±0.51    & 54.07±0.52           & 80.73±0.86   & 59.60±0.71          & 69.91±0.93     & 48.88±0.94            & 90.51±1.20   & 52.01±2.17           & 80.79±1.99     & 26.07±1.43            \\
+LPA     & 78.68±0.49            & 53.88±1.10           & 81.94±0.51           & 62.96±0.75          & 69.98±1.02             & 49.51±1.11            & 90.42±1.31           & 55.06±1.42           & 81.32±2.03             & 27.67±2.36            \\
+DPM-SNC & 81.63±0.68    & 60.06±0.82   & \textbf{82.47±0.69}   & 62.32±1.27  & 72.19±0.86     & 52.11±0.67    & \textbf{91.45±1.11}   & 53.63±2.15   & 83.04±1.93     & 29.39±3.09    \\
+\textbf{\ours}    & \textbf{83.16±0.67}    & \textbf{60.08±1.04}   & 82.27±0.85   & \textbf{63.93±0.80}  & \textbf{73.00±1.04}     & \textbf{52.81±0.81}    & 91.30±1.35   & \textbf{56.04±2.29}   & \textbf{83.18±1.67}     & \textbf{30.51±2.86}    \\ \midrule
GAT      & 77.42±1.16    & 53.90±1.08           & 81.02±0.85   & 59.32±1.33          & 69.65±1.37     & 49.11±1.24            & 90.49±1.08   & 51.90±2.83           & 82.36±1.98     & 28.16±3.29            \\
+LPA     & 78.87±0.70            & 56.16±1.51           & 81.40±0.78           & 60.55±1.51          & 70.62±0.91             & 49.89±0.60            & 90.92±1.74           & 55.45±2.41           & 83.05±1.90             & 28.94±2.66            \\
+DPM-SNC & 81.90±1.25    & 60.37±1.36   & \textbf{82.88±0.57}   & \textbf{64.03±1.02}  & \textbf{74.11±0.90}     & 53.16±0.48    & 90.63±2.24   & 53.52±2.59   & 83.89±1.91     & 28.91±2.69    \\
+\textbf{\ours}    & \textbf{82.15±0.68}           & \textbf{60.56±0.97}           & 82.29±0.52           & 62.74±0.81          & 73.23±0.91             & \textbf{53.28±0.71}            & \textbf{91.59±1.07}           & \textbf{57.32±1.42}           & \textbf{84.18±1.96}            & \textbf{29.16±1.30}            \\ \bottomrule
\end{tabular}
}
\end{table*}

\begin{table}[t]
\centering
\caption{Node-level accuracies on heterophilic graphs.}
\label{tab:heter_results}
\begin{tabular}{lccc}
\toprule
 & \textbf{Roman-empire} & \textbf{Amazon-ratings} & \textbf{Minesweeper} \\
\midrule
H$_2$GCN & 60.11±0.52 & 36.47±0.23 & 32.81±0.93 \\
CPGNN & 63.96±0.62 & 39.79±0.27 & 52.03±5.46 \\
GPR-GNN & 64.85±0.27 & 44.88±0.34 & 50.08±0.75 \\
FSGNN & 79.92±0.56 & 52.74±0.83 & 50.02±0.70 \\
GloGNN & 59.63±0.69 & 36.89±0.31 & 51.08±1.23 \\
FAGCN & 65.22±0.46 & 41.25±0.27 & 52.13±0.56 \\
GBK-GNN & 74.57±0.47 & 45.98±0.34 & 48.03±0.66 \\
JacobiConv & 71.14±0.42 & 43.55±0.48 & 89.66±0.40 \\
\midrule
GCN & 73.69±0.74 & 48.70±0.63 & 89.75±0.52 \\
SAGE & 85.74±0.67 & 53.63±0.39 & 93.51±0.31 \\
GAT & 80.87±0.30 & 49.09±0.63 & 91.02±0.65 \\
GAT-sep & 88.75±0.41 & 52.70±0.62 & 93.91±0.35 \\
GT & 86.51±0.73 & 51.17±0.66 & 91.65±0.76 \\
GT-sep & 87.32±0.39 & 52.18±0.80 & 92.29±0.47 \\
\midrule
LGD & 88.67±0.08 & 52.17±0.17 & 89.86±0.30 \\
DPM-SNC & 89.52±0.46 & \textbf{54.66±0.39} & 94.19±0.55 \\
\textbf{\ours} & \textbf{89.88±0.38} & 53.92±0.61 & \textbf{94.55±0.58} \\
\bottomrule
\end{tabular}
\end{table}

\begin{table}[bthp]
\centering
\caption{Statistics of homophilic and heterophilic graphs.}
\label{tab:datasets}
\begin{tabular}{lllll}
\hline
          & \#Nodes & \#Edges & \#Features & \#Classes \\ \hline
Cora      & \num{2708}    & \num{5278}    & \num{1433}       & \num{7}         \\
CiteSeer  & \num{3327}    & \num{4522}    & \num{3703}       & \num{6}         \\
PubMed    & \num{19717}   & \num{44324}   & \num{500}        & \num{3}         \\
Computer & \num{13381}   & \num{34493}  & \num{767}        & \num{10}        \\
Photo     & \num{7487}    & \num{119043}  & \num{745}        & \num{8}         \\
Roman-empire & \num{22662}    & \num{32927}   & \num{300}       & \num{18}         \\
Amazon-ratings  & \num{24492}    & \num{93050}   & \num{300}    & \num{5}         \\
Minesweeper   & \num{10000}    & \num{39402}   & \num{7}        & \num{2}   \\
\hline
\end{tabular}%
\end{table}

\subsection{Studies on homophilic graphs}
\label{subsec:homo_and_heter}
Our experiments focus on structured label prediction, assessing \ours's ability to enhance both node-level accuracy and subgraph-level consistency on homophilic graphs, where labels are smoothly distributed.

\noindent\textbf{Datasets}.
Following DPM-SNC~\cite{DPM-SNC}, we conduct experiments on eight widely-used datasets,  including five homophilic graphs Cora, CiteSeer, PubMed~{\cite{citation1,citation2}}, Computers, Photo~\cite{cop1,cop2}.
A detailed description of these datasets is provided in Table~\ref{tab:datasets}.

\noindent\textbf{Protocol}.
For all homophilic graphs, we randomly sample 20 nodes per class for training, 30 nodes per class for validation, and use the remaining nodes for testing. We evaluate all GNN-based methods with GCN~\cite{GCN} and GAT~\cite{GAT} as backbone architectures, and report both node classification accuracy and subgraph accuracy. Here, subgraph accuracy measures the proportion of nodes whose own prediction and the predictions of all their immediate neighbors are simultaneously correct, thus reflecting the consistency of local label structures.

\noindent\textbf{Implementation details}.
For the simple model, we adopt the same GNN backbone as used in \ours. All simple models are trained for 500 epochs, and the best checkpoint is selected based on the validation accuracy. We perform hyperparameter optimization for both the simple model and \ours, with the learning rate in \{0.001, 0.005, 0.01\}, weight decay in \{0.001, 0.005\}, and the priority queue temperature $\tau$ in \{0.01, 0.1, 0.5\}. All models use a two-layer architecture. When using GCN as the backbone, the hidden size is set to 64. For GAT, we employ 8 attention heads, each with a hidden size of 8. The optimal hyperparameters are selected based on validation performance. For both \ours and DPM-SNC, the number of diffusion steps $T$ is fixed at 80 and the size of priority queue $S$ is set to 100.

\noindent\textbf{Baselines}.
We compare our method with representative GNNs, including GCN~\cite{GCN} and GAT~\cite{GAT}, as well as label propagation-based methods such as LP~\cite{LP} and LPA~\cite{LPA}. In addition, we consider the method with structure prediction, i.e., DPM-SNC~\cite{DPM-SNC}, and the SOTA diffusion-based method, LGD~\cite{LGD}.

\noindent\textbf{Results and Analysis}.
Table~\ref{tab:homo_results} shows that \ours outperforms or matches all baselines across both evaluation metrics, using either GCN or GAT as the backbone. These results demonstrate the effectiveness of \ours on homophilic graphs.
Compared to GCN, \ours achieves an average improvement of 3.1\% in node-level accuracy and 10.3\% in subgraph-level accuracy. When using GAT, the gains are 3.2\% and 8.1\%, respectively. These consistent improvements underscore the effectiveness of \ours in enhancing both individual predictions and structural consistency.
Notably, when using GCN as the backbone, \ours consistently achieves the best subgraph-level accuracy across all datasets. With GAT, \ours achieves the best subgraph-level accuracy on five out of six datasets, with only a slight drop on Cora.
Compared to DPM-SNC~\cite{DPM-SNC}, \ours generally achieves better performance, which can be attributed to its discrete diffusion formulation that models label distributions directly in their natural discrete space, avoiding the semantic mismatch introduced by learning in continuous latent spaces for inherently discrete classification targets.

In addition, the relatively poor performance of LP highlights the limitation of non-parametric methods: LP is unable to model complex label dependencies and does not exploit node feature information, resulting in inferior accuracy at both node and subgraph levels. Finally, it is worth noting that the performance gap observed for LGD stems partly from our adoption of more challenging and realistic data splits. Unlike those used in its original paper, our splits better reflect practical deployment scenarios, yet impose additional difficulty for methods like LGD, leading to a noticeable drop in performance on homophilic graphs.

\subsection{Studies on heterophilic graphs}
We further evaluate \ours on heterophilic graphs, where traditional label propagation methods are limited by their reliance on homophily assumptions.

\noindent\textbf{Datasets}.
We conduct experiments on three heterophilic graphs, namely Roman Empire, Amazon Ratings, and Minesweeper~\cite{heter}, to evaluate the performance of \ours under heterophily. The dataset statistics are provided in Table~\ref{tab:datasets}.

\noindent\textbf{Protocol}.
For heterophilic graphs, we adopt the default 50\%/25\%/25\% split used in~\cite{heter} for training, validation, and testing, with each random seed corresponding to a different split. Following DPM-SNC~\cite{DPM-SNC}, we adopt GAT-sep~\cite{heter} as the backbone and report node classification accuracy.

\noindent\textbf{Implementation details}. 
We perform hyperparameter optimization for both the simple model (i.e., GAT-sep~\cite{heter}) and \ours, with the learning rate selected from $\{0.0001, 0.0003, 0.001\}$, weight decay from $\{0, 0.001, 0.005\}$, and the priority queue sampling temperature from {0.01, 0.25, 0.5}. All other hyperparameter settings are consistent with those used in experiments on homophilic graphs.

\noindent\textbf{Baselines}.
We compare \ours with six general-purpose GNNs: GCN~\cite{GCN}, GraphSAGE~\cite{GraphSAGE}, GAT~\cite{GAT}, GAT-sep~\cite{heter}, GT~\cite{GraphTransformer}, and GT-sep~\cite{heter}. We further include comparisons with eight heterophily-oriented GNNs, including H2GCN~\cite{H2GCN}, CPGNN~\cite{H2GCN}, GPR-GNN~\cite{GPRGNN}, FSGNN~\cite{FSGNN}, GloGNN~\cite{GloGNN}, FAGCN~\cite{FAGCN}, GBK-GNN~\cite{GBKGNN}, and JacobiConv~\cite{Jacob}. Particularly, we consider two representative diffusion-based methods---DPM-SNC~\cite{DPM-SNC} and LGD~\cite{LGD}.

\noindent\textbf{Results and Analysis}.
As shown in Table 2, our method consistently achieves top-tier performance across all datasets. In particular, \ours achieves the best accuracy on Roman-Empire , Minesweeper, and competitive performance on Amazon-Ratings, closely following DPM-SNC.
Compared to the backbone GNN, i.e., GAT-sep~\cite{heter}, our method yields consistent improvements, achieving an average performance gain of 1.9\% across the three heterophilic datasets.
Notably, our method also outperforms recent heterophily-oriented models including H2GCN~\cite{H2GCN}, GPR-GNN~\cite{GPRGNN}, and FSGNN~\cite{FSGNN}, demonstrating the ability of \ours to capture long-range label dependencies and perform accurate structured prediction, even when local neighborhood information is misaligned with label distributions.

\begin{table}[th]
\caption{Accuracies on large graphs at both node-level (N-Acc.) and subgraph-level (Sub-Acc.).}
\label{tab:large-graph}
\begin{tabular}{@{}ccccc@{}}
\toprule
         & \multicolumn{2}{c}{Arxiv} & \multicolumn{2}{c}{Products}  \\ 
Method   & N-Acc.       & Sub-Acc.    & N-Acc.      & Sub-Acc.    \\ \midrule
LP       & 66.06±0.0            & 24.86±0.0           & 72.32±0.0           & 39.85±0.0         \\
PTA      & 56.43±0.35            & 7.20±0.14           & 71.35±0.14           & 41.20±0.18           \\ 
\midrule
LGD & 69.68±0.07     & 25.43±0.59     & OOM   & OOM \\

\midrule
SAGE      & 70.53±0.19    & 23.65±0.97           & 78.93±0.31   & 42.62±0.15         \\
+LPA     & 70.98±0.54     & 26.18±0.75           & 78.96±0.28  &  43.21±0.22                \\
+DPM-SNC & 65.28±0.78    & 38.34±0.81   & OOM   & OOM \\
+\textbf{\ours}    & \textbf{72.32±0.17}    & \textbf{44.43±0.10}   & \textbf{79.28±0.35}   & \textbf{45.91±0.24} \\ \midrule
\end{tabular}
\end{table}

\begin{table}[htbp]
\centering
\caption{Statistics of large graphs.}
\label{tab:ogb datasets}
\begin{tabular}{lll}
\hline
             & ogbn-arxiv & ogbn-products \\ \hline
\#Nodes      & \num{169343}     & \num{2449029}       \\
\#Train      & \num{90941}      & \num{196615}        \\
\#Validation & \num{29799}      & \num{39323}         \\
\#Test       & \num{48603}      & \num{2213091}       \\
\#Edges      & \num{1166243}    & \num{123718280}     \\
\#Features   & \num{128}        & \num{100}           \\
\#Classes    & \num{40}         & \num{47}            \\ \hline
\end{tabular}%
\end{table}

\subsection{Studies on large graphs}
\label{subsec:ogbexp}
To evaluate scalability, we test \ours on large-scale graphs to verify whether it maintains strong performance as the graph size and label space grow.

\noindent\textbf{Datasets}
This experiment aims to verify the effectiveness of \ours on large graphs, such as ogbn-arxiv and ogbn-products. Statistics about them can be found in Table~\ref{tab:ogb datasets}.

\noindent\textbf{Protocol}.
We use the default data splits provided by OGB~\cite{OGB} for both graphs and adopt GraphSAGE as the backbone for all GNN-based methods. In our experiments, we report both node classification accuracy and subgraph accuracy as evaluation metrics.

\noindent\textbf{Implementation details}.
We adopt GraphSAGE~\cite{GraphSAGE} as the backbone for the simple model. The hidden layer size is set to 256, while other hyperparameter settings follow those used for experiments on homophilic graphs. For efficiency considerations, we reduce the number of diffusion steps to 40 when evaluating on the Products dataset, and set the training epochs to 60 for the simple model and 200 for \ours.

\noindent\textbf{Results and Analysis}.
As shown in Table~\ref{tab:large-graph}, \ours consistently achieves superior performance on large graphs in terms of both metrics. Compared to the simple model, i.e., GraphSAGE, \ours delivers substantial improvements, with gains of 1.5\% in node-level accuracy and 47.8\% in subgraph-level accuracy, demonstrating its effectiveness in producing structured predictions.

In particular, while existing methods such as LGD~\cite{LGD} and DPM-SNC~\cite{DPM-SNC} encounter out-of-memory (OOM) errors when processing the Products dataset, our proposed approach demonstrates superior scalability by successfully handling this large graph while maintaining competitive performance. This empirical evidence substantiates the time complexity analysis presented in Sec.~\ref{subsec:analysis}, demonstrating that REDISC effectively scales to large graphs and thereby affirming its practicality for deployment on complex real-world graph data.

\subsection{Ablation Study}
\label{subsec:ablation}
To assess the contribution of each innovative component in \ours, we conduct ablation studies focusing on three key aspects: (1) the priority queue, designed to select higher-quality pseudo-labels and stabilize training; (2) the labeled-first inference strategy, which leverages ground-truth labels early in the denoising process to guide label dependencies; and (3) the time-aware GNN layer, intended to adaptively incorporate timestep information for more effective feature fusion. We adopt GCN~\cite{GCN} and GAT-sep~\cite{heter} as backbones for homophilic and heterophilic graphs, respectively, to ensure architectural alignment with graph properties. All experiments are performed on the Cora~\cite{citation2}, Computers, Photo~\cite{cop1, cop2}, Roman-empire, and Minesweeper~\cite{heter}, using node classification accuracy as the evaluation metric. These controlled experiments validate the importance of each design choice and offer insights into their individual and combined effects on overall performance.

\noindent \textbf{Priority queue.}
To evaluate the impact of the priority queue, we compare full \ours to a variant that removes this component and instead uniformly selects cached pseudo-labels. As shown in Table~\ref{tab:priority_queue}, incorporating the priority queue consistently leads to improved performance across all datasets, resulting in an average improvement of 0.31\%.
Notably, the improvements are more pronounced on heterophilic graphs such as Roman-empire and Minesweeper, with accuracy increases of 0.54\% and 0.45\%, respectively, which suggests that quality-aware pseudo-label selection is particularly beneficial when label correlations are weak or even reversed.
By prioritizing pseudo-labels that yield higher validation accuracy, our masked diffusion model receives more reliable observations in the M-step, which tends to produce a more useful model update.

\begin{table}[ht]
\centering
\caption{Accuracy comparison with (w/) and without (w/o) the priority queue.}

\begin{tabular}{cccc}
\toprule
\textbf{Dataset} & \textbf{w/} & \textbf{w/o} & \textbf{$\Delta$} $\uparrow$ \\
\midrule
Cora        & \textbf{82.27±0.85}  & 82.00±0.69 & 0.27 \\
Computer      & \textbf{83.18±1.96}  & 82.08±1.95 & 0.10 \\
Photo         & \textbf{91.30±1.35} & 91.11±1.20 & 0.19 \\
Roman-empire   & \textbf{89.88±0.38} & 89.34±0.78 & 0.54 \\
Minesweeper   & \textbf{94.55±0.58} & 94.10±0.87 & 0.45 \\
\bottomrule
\end{tabular}
\label{tab:priority_queue}
\end{table}

\noindent \textbf{Labeled-First inference.}
To evaluate the importance of the labeled-first inference strategy, we compare \ours with and without inferring for labeled nodes at first. As shown in Table~\ref{tab:label_first}, incorporating the labeled-first inference strategy yields consistent and notable improvements across all datasets.
Overall, the labeled-first inference strategy results in an average improvement of 1.22\% in node classification accuracy, demonstrating its effectiveness in leveraging reliable supervision to enhance the denoising process and achieve more accurate structured predictions. In particular, we observe substantial performance gains on Minesweeper, where accuracy increases by 4.98\% with this strategy, demonstrating its effectiveness in modeling label dependencies on heterophilic graphs.

\begin{table}[ht]
\centering
\caption{Accuracy comparison with (w/) and without (w/o) the labeled-first inference strategy.}

\begin{tabular}{cccc}
\toprule
\textbf{Dataset} & \textbf{w/} & \textbf{w/o} & \textbf{$\Delta$} $\uparrow$ \\
\midrule
Cora        & \textbf{82.27±0.85}  & 82.12±0.71 & 0.15 \\
Computer      & \textbf{83.18±1.96}  & 81.91±1.74 & 0.73 \\
Photo         & \textbf{91.30±1.35} & 91.08±1.00 & 0.22 \\
Roman-empire   & \textbf{89.88±0.38} & 89.85±0.41 & 0.03 \\
Minesweeper   & \textbf{94.55±0.58} & 89.57±0.57 & 4.98 \\
\bottomrule
\end{tabular}
\label{tab:label_first}
\end{table}

\noindent \textbf{Time-aware GNN Layer.}
To assess the impact of our architectural modifications, we compare \ours using time-aware GNN layers that incorporate timestep information through a scaling mechanism with its counterpart employing vanilla GNN layers across multiple datasets. As presented in Table~\ref{tab:gnn_arch}, \ours with the time-aware GNN layers consistently outperforms the vanilla GNN layers, demonstrating the benefit of the proposed design. In particular, the improvements are evident on both homophilic graphs (e.g., Cora, Computer, Photo) and heterophilic graphs (e.g., Roman-empire and Minesweeper). The most substantial gains appear on the Computer and Minesweeper datasets, with increases of 1.27\% and 0.82\% respectively. These results indicate that the proposed timestep-aware scaling mechanism enables more effective fusion of node features and label embeddings, allowing the model to adaptively emphasize relevant information based on the temporal context. Although the improvements on Cora and Roman-empire are relatively modest, the consistent positive trend across all datasets underscores the generalizability of the proposed architectural modification. Overall, these findings demonstrate the advantage of employing a learnable scaling strategy over direct addition for incorporating temporal information into the denoising process.

\begin{table}[ht]
\centering
\caption{Performance comparison of \ours equipped with time-aware GNN layers versus vanilla GNN layers across datasets. $\Delta$ indicates the performance difference between the two configurations.}

\begin{tabular}{cccc}
\toprule
\textbf{Dataset} & \textbf{Time-aware} & \textbf{Vanilla}  & \textbf{$\Delta$} $\uparrow$ \\
\midrule
Cora        & \textbf{82.27±0.85}  & 82.21±0.67  & 0.06 \\
Computer      & \textbf{83.18±1.96} & 81.91±1.74   & 1.27 \\
Photo        & \textbf{91.30±1.35}  & 90.70±0.92 & 0.60 \\
Roman-empire   & \textbf{89.88±0.38} & 89.78±0.44  & 0.10 \\
Minesweeper   & \textbf{94.55±0.58} & 93.73±0.87  & 0.82 \\
\bottomrule
\end{tabular}
\label{tab:gnn_arch}
\end{table}

\subsection{Sensitivity analysis}
\label{subsec:sensitivity}

\begin{figure}[ht]
  \centering
  \begin{minipage}[t]{0.48\linewidth}
    \centering
    \includegraphics[width=\linewidth]{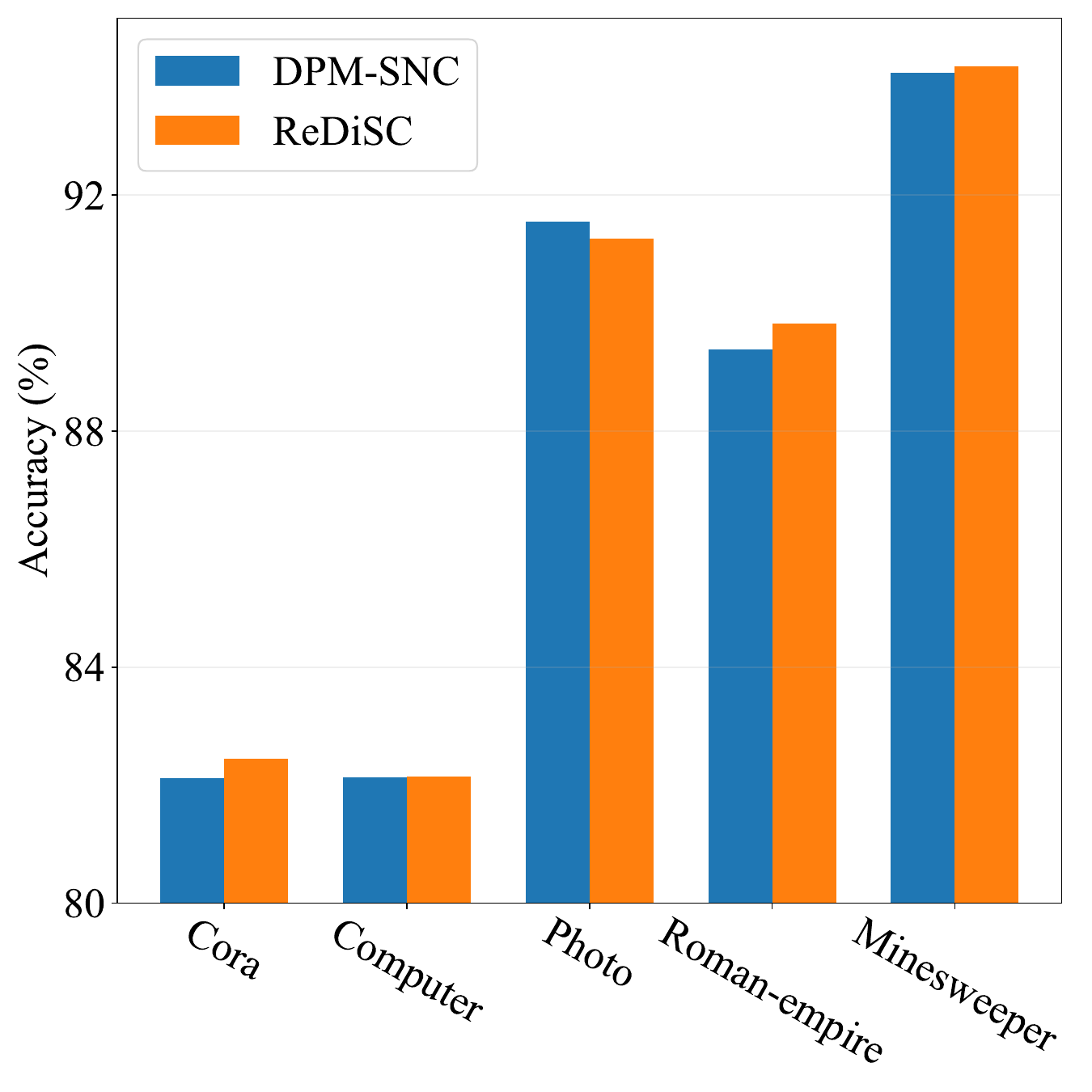}
    \label{fig:step40}
  \end{minipage}
  \hfill
  \begin{minipage}[t]{0.48\linewidth}
    \centering
    \includegraphics[width=\linewidth]{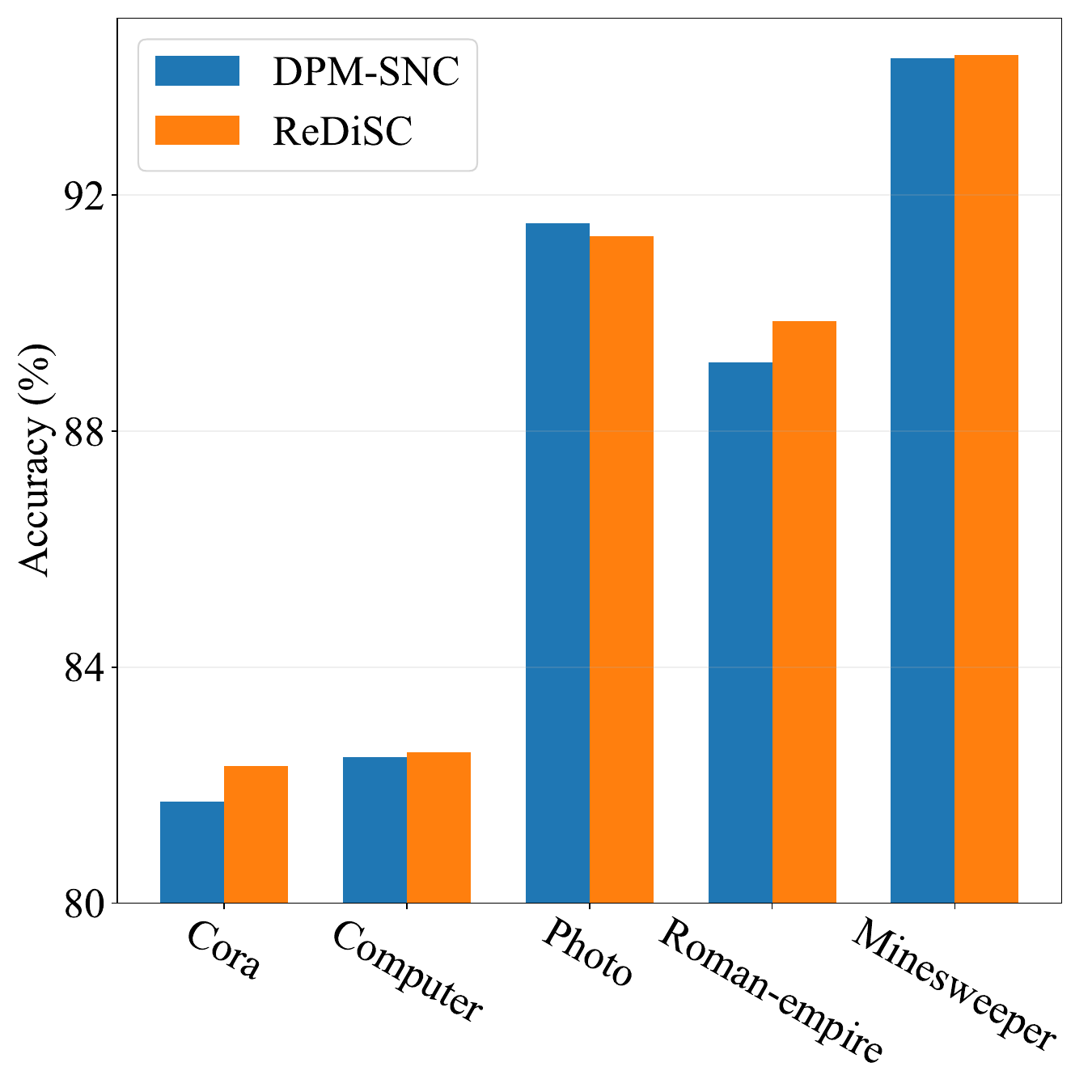}
    \label{fig:step60}
  \end{minipage}

  \caption{Accuracy comparison with $\mathrm{T}=40$ (left) and $\mathrm{T}=60$ (right) across various datasets.}
  \label{fig:diffusion_steps}
\end{figure}

To examine how effectively the model can denoise and recover label signals under limited diffusion steps, which is important for computational efficiency, we conduct experiments with smaller values of $\mathrm{T}$, specifically $\mathrm{T} = 40$ and $\mathrm{T} = 60$. As shown in Fig.~\ref{fig:diffusion_steps}, \ours outperforms DPM-SNC on most datasets under both $\mathrm{T}=40$ and $\mathrm{T}=60$, with particularly notable gains on the Cora and Roman-Empire datasets. This demonstrates that \ours maintains strong predictive accuracy even with fewer denoising steps, highlighting its robustness to the choice of $\mathrm{T}$.
This advantage under shorter diffusion schedules likely arises from the fact that \ours directly models label dependencies in a discrete space, which aligns well with the intrinsic discrete nature of node classification and enables more efficient label recovery than methods operating in continuous latent spaces. These findings suggest that practitioners can safely adopt fewer diffusion steps (e.g., $\mathrm{T}=40$ or $60$) to achieve substantial computational savings without significant loss of accuracy, which is especially beneficial for large-scale graphs.

\section{Discussion}
\label{sec:discussion}
This work introduces \ours, a reparameterized masked diffusion model that brings discrete-domain diffusion to structured node classification. By aligning the generative process with the inherently discrete nature of node labels, \ours bridges the gap between expressive diffusion models and practical graph tasks. Our theoretical and empirical results highlight its advantages in scalability, semantic alignment, and interpretability, positioning it as a strong alternative to continuous-domain diffusion and classical ``GNN+LP'' hybrids.
Beyond node classification, this framework opens new avenues for discrete generative modeling on graphs and other structured data.
Despite these strengths, \ours still has limitations. The variational EM framework introduces approximation gaps that may limit performance on graphs with highly complex label dependencies. Additionally, our current design focuses on transductive scenarios; extending the approach to fully inductive or dynamic settings remains non-trivial.
Future work includes extending \ours to dynamic graphs, investigating principled active sampling strategies for efficiency, and distilling \ours to a more efficient one-shot classifier. We hope our findings inspire further research at the intersection of discrete diffusion, structured prediction, and scalable graph learning.

\bibliographystyle{abbrv}
\bibliography{refs}
%



\end{document}